%% file: main.tex
\documentclass{article}

\usepackage{PRIMEarxiv}

\usepackage[utf8]{inputenc} 
\usepackage[T1]{fontenc}    
\usepackage{hyperref}       
\usepackage{url}            
\usepackage{booktabs}       
\usepackage{amsfonts}       
\usepackage{nicefrac}       
\usepackage[linesnumbered,lined]{algorithm2e}
\usepackage{amsthm, amssymb,amsmath}
\usepackage{float}

\usepackage{microtype}      
\usepackage{lipsum}
\usepackage{fancyhdr}       
\usepackage{graphicx}       

\newtheorem{theorem}{Theorem}
\newtheorem{definition}{Definition}
\newtheorem{lemma}[theorem]{Lemma}

\newtheorem{corollary}[theorem]{Corollary}

\newtheorem{theorema}{Theorem}[section]

\def\~#1{\mathbb{#1}}
\def\*#1{\mathbf{#1}}
\def\@#1{\mathcal{#1}}

\RestyleAlgo{ruled}

\title{Accurate Coresets for Latent Variable Models and Regularized Regression}

\author{
  Sanskar Ranjan \\
  IIIT Delhi \\
  \texttt{sanskar21096@iiitd.ac.in}
   \And
  Supratim Shit \\
  IIIT Delhi \\
  \texttt{supratim@iiitd.ac.in}
}

\begin{document}
\maketitle

\begin{abstract}
Accurate coresets are a weighted subset of the original dataset, ensuring a model trained on the accurate coreset maintains the same level of accuracy as a model trained on the full dataset. Primarily, these coresets have been studied for a limited range of machine learning models. In this paper, we introduce a unified framework for constructing accurate coresets. Using this framework, we present accurate coreset construction algorithms for general problems, including a wide range of latent variable model problems and $\ell_p$-regularized $\ell_p$-regression. For latent variable models, our coreset size is $O\left(\mathrm{poly}(k)\right)$, where $k$ is the number of latent variables. For $\ell_p$-regularized $\ell_p$-regression, our algorithm captures the reduction of model complexity due to regularization, resulting in a coreset whose size is always smaller than $d^{p}$ for a regularization parameter $\lambda > 0$. Here, $d$ is the dimension of the input points. This inherently improves the size of the accurate coreset for ridge regression. We substantiate our theoretical findings with extensive experimental evaluations on real datasets.
\end{abstract}

\keywords{Accurate Coreset \and Regularized Regression \and Latent Variable Models \and Tensor Decomposition}

\input{section/introduction}

\input{section/preliminary}

\input{section/framework}

\input{section/experiment}


\bibliographystyle{unsrt}  
\bibliography{references}  

\appendix
\input{section/appendix}

\end{document}

%% file: section/introduction.tex
\section{Introduction}\label{sec:intro}
In the current era, coresets are very useful due to their provable theoretical guarantees. The most widely used coresets suffer from a loss in accuracy compared to the accuracy obtained by training a model on the full data. Accurate coresets are very special due to their powerful theoretical guarantees. These are not only weighted subsets of the full data, but they also retain the exact loss that would have been incurred from the full data. 

Let $\*D$ be a dataset, $\*Q$ be a query space and a loss function $f: \*D \times \*Q \to \~R_{\geq 0}$. For any $\*q \in \*Q$, let $f(\*D,\*q) = \sum_{\*d \in \*D}f(\*d,\*q)$. The goal of any model training algorithm is to compute a $\*q^{*} \in \*Q$, such that $f(\*D,\*q^*)$ is the smallest among all possible $\*q \in \*Q$. An accurate coreset for this problem is a weighted set $\*D_c$ and its corresponding weight function $\*w: \*D_c \to \~R_{>0}$ such that, for every $\*q \in \*Q$ we have,

\begin{equation}
    f(\*D_c,\*q) = f(\*D,\*q). \nonumber
\end{equation}

Here, $f(\*D_c,\*q) = \sum_{\*d \in \*D_c}w(\*d)f(\*d,\*q)$. As an accurate coreset retains the exact loss functions for every $\*q \in \*Q$, so training a model on the accurate coreset return the exact model that could have been trained on the full data. The algorithms for constructing accurate coresets is limited to very specific and sometime trivial problems. 

In this paper we present algorithm for constructing accurate coresets for a wide range of problems in latent variable modeling and regularized $\ell_p$ regression. 


Our main contributions in this paper are as follows.
\begin{itemize}
    \item We present a unified framework (Algorithm \ref{alg:framework}) for constructing accurate coresets for any machine learning algorithm. In this framework, we introduce $\mathrm{Kernelization}$ (Definition \ref{def:Kernelization}) of the dataset, which maps every input point to a high dimensional space while retaining the loss on the full data for any query with the points in this high dimensional space. One of the main challenges here is designing such a $\mathrm{Kernelization}$ function for a dataset for any ML model.
    \item We present Algorithm \ref{alg:ridge} for constructing an accurate coreset for Ridge Regression. Our algorithm uses a novel $\mathrm{Kernelization}$ (Lemma \ref{lem:RidgeRegression}) that supports our analysis of quantifying the size of the coreset. We show that the size of the accurate coreset decreases with a reduction in model complexity due to some regularization parameter $\lambda > 0$. Our coreset size depends on the statistical dimension, which is always smaller than $d^{2}$ for $\lambda > 0$ (Theorem \ref{thm:ridge}).
    \item We further present Algorithm \ref{alg:lpReguRegression} that generalize our ridge regression result to $\ell_p$ regularized $\ell_p$ regression for any even valued $p$. It returns an accurate coreset whose size is smaller than $d^{p} + 1$ (Theorem \ref{thm:lpReguRegression}). 
    \item We also provide Algorithm \ref{alg:lvm} for constructing accurate coresets for a wide range of problems in latent variable models such as Gaussian Mixture Models, Hidden Markov Models, Latent Dirichlet Allocation etc. Our coreset selects at most $O(k^3)$ samples (Theorem \ref{thm:lvm}). The coreset particularly preserves tensor contraction (Corollary \ref{cor:TensorContraction}) which is useful in recovering the exact latent variables from the coreset, that could have been recovered from the full data. 
    \item Finally, we conducted extensive empirical evaluations on multiple real datasets and show that the results support our theoretical claims. We demonstrate that, in practice, our coresets for latent variable models are less than $1\%$ of the full data, due to which there is significant reduction in the training time. 
\end{itemize}

%% file: section/preliminary.tex
\section{Preliminary}\label{sec:Prelim}

\subsection{Notations}\label{sec:notations}
A scalar is denoted by a simple lower or uppercase letter, e.g., $p$ or $M$ while a vector is denoted by a boldface lowercase letter, e.g., $\*x$. By default, all vectors are considered as column vectors unless specified otherwise. Matrices are denoted by boldface upper case letters, e.g., $\*X$. Specifically, $\*X$ denotes an $n\times d$ matrix where $n$ is the number of points and in $\~R^d$ feature space. Normally, we represent the $i^{th}$ row of $\*X$ as $\*x_{i}^{T}$ and $j^{th}$ column is $\*x_{j}$ unless stated otherwise. $\*1$ and $\*0$ are vectors with all indices having value $1$ and $0$ respectively.
We denote 2-norm of a vector and a matrix (or spectral norm) as $\|\*x\|_2$ and $\|\*X\|_{(2)} := \max_\*y \frac{\|\*X\*y\|_2^2}{\|\*y\|_2^2}$ respectively. The square of the Frobenius norm is defined as $\|\*X\|_F^2 := \sum_{i,j} X_{i,j}^2$. For any $p \in (0, \infty)$ the $\ell_p$ norm of a matrix is defined as $\|\*X\|_p := \left(\sum_{i,j} X_{i,j}^p\right)^{1/p}$. A Tensor is a higher-order matrix. It is denoted by a bold calligraphy letter e.g., $\@T$. Given a set of $d-$dimensional vectors $\{\*x_1, \ldots, \*x_n\}$, a $p$-order tensor is defined as $ \@T = \sum_i^n \*x_i \otimes^{p}$ i.e. the sum of the $p$-th order outer product of each of the vectors. It is also known as $p$-order moment. Notice, $\forall i_{1},i_{2},_{\cdots},i_{p}$ tensor $\@T$ satisfies that $\@T_{i_{1},i_{2},_{\cdots},i_{p}} = \@T_{i_{2},i_{1},_{\cdots},i_{p}} = _{\cdots} = \@T_{i_{p},i_{p-1},_{\cdots},i_{1}}$, i.e. same value for all possible permutations of $(i_1, i_2, _{\cdots}, i_p)$. We define tensor contraction operation, as $\@T(\*w,\ldots,\*w) = \sum_{i=1}^{n}(\*x_i^T\*w)^p$, where $\*w \in \~R^{d}$. For a matrix $\*M \in \~R^{d \times d}$, $\mathrm{vec}(\*M) \in \~R^{d^2}$. The $\mathrm{vec}()$ represents the vector form of the matrix $\*M$. It also extends to Tensor. For a given vector $\*x \in \~R^d$, $\*M := \mathrm{diag}(\*x)$ is a diagonal matrix of size $d \times d$ such that for every $i \in [d]$, $M_{i,i} = x_i$. 

\subsection{Related Work}\label{sec:related}
Given a dataset and a loss function to optimize, its coreset is a weighted subsample of the dataset that ensures certain theoretical guarantees. Typically, there are three types of coresets.
\begin{itemize}
    \item {\em Indeterministic Coresets} are the most commonly used coresets in practice. These coresets ensure that the loss on them well approximates the loss on the complete dataset for every possible model with some probability \cite{harpeled2004on}. Hence, the coreset's guarantees suffers from failure probability. Typically, these coresets are constructed based on a distribution defined over the points in the dataset based on the loss function that needs to be optimized \cite{feldman2011unified}. These are extensively studied for multiple problems such as regression \cite{dasgupta2009sampling, avron2016sharper, chhaya2020coresets,woodruff2023online}, classification \cite{munteanu2018coresets, tukan2020coresets, mai2021coresets}, clustering \cite{feldman2013turning, cohenaddad2021new, cohenaddad2022towards} and mixture model \cite{feldman2011scalable, lucic2018training, chhaya2020streaming}.
    \item {\em Deterministic Coresets} ensures that the loss on the complete dataset can be well approximated by the loss on the coreset points with probability $1$. Deterministic coresets were first introduced in the seminal BSS algorithm \cite{batson2009twice}. The paper presented a graph sparsification algorithm that deterministically preserved the spectral properties of the original graph. They use an iterative selection method for constructing the subgraph (i.e., coreset). This motivated multiple results in the machine learning community, such as clustering \cite{cohen2015dimensionality, shit2022online}, fast matrix multiplication \cite{cohen2015optimal} and regression \cite{boutsidis2013near, kacham2020optimal}.
    \item {\em Accurate Coresets} are also deterministic coresets, with additional features that the loss on the complete dataset is exactly equal to the loss on the coreset. They use a geometric property of the dataset in a space defined by the loss function and the space in which the dataset is present. Due to this, these coresets are only restricted to a limited problems in the machine learning such as $1$-mean, $1$-segment, and linear regression \cite{maalouf2019fast, jubran2021overview}. \cite{maalouf2019fast} shows the size of accurate coreset for ridge regression is $O((d+1)^2)$ where input points are $\~R^d$ and $\lambda > 0$ is some regularization parameter. In one of our contribution we have improved this result. 
\end{itemize}

%% file: section/framework.tex
\section{Accurate Coreset}\label{sec:accurate}
The construction of an accurate coreset exploits properties from computational geometry. For a given optimization function, it maps the original points of the full dataset into another space. Then, it considers the convex hull of these point sets, where the goal is to identify a point in the hull that can be used to compute the loss of the full data. A machine learning model whose loss can be realized as inner products between points and model parameters in the high dimensional space can have accurate coresets. The accurate coreset construction uses the following well known result from computational geometry. 
\begin{theorem}{\label{thm:caratheodory}}
    Let $\*P$ be a set of $n$ points in $\~R^d$ and it spans $k$-dimensional space. If $\*x$ is a point inside the convex hull of $\*P$, then $\*x$ is also in the convex hull of at most $k+1$ weighted points in $\*P$.
\end{theorem}

The above theorem is well known as Caratheodory's Theorem \cite{cook1972caratheodory, caratheodory1907variabilitatsbereich} and is a classical result in computational geometry. It states that any point $\*x$ that can be represented as a convex combination of points in $\*P$ can also be represented as the convex combination of at most $k+1$ weighted points from $\*P$, where $P$ spans a $k$-dimensional space. For our problems, one of the important tasks is to identify such a point that can be used to accurately preserve the loss of the full dataset for any query. If this point is a convex combination of the dataset i.e., lies inside its convex hull, then it can be represented using a convex combination of at most $k+1$ points. These points represent the accurate coreset of the dataset for the problem. 

\subsection{Unified Framework for Accurate Coreset}
In this section, we describe a unified framework for constructing an accurate coreset for general problems. Let $\*X$ be a dataset of $n$ points. In the case of unsupervised models, every $\*x \in \*X$ is a point in $\~R^d$, and in the case of supervised learning, every $\*x \in \*X$ represents both point and its label. Let $f:\*X \times \*Q \to \~R_{\geq 0}$ be a function that acts upon points $\*x \in \*X$ and queries $\*q \in \*Q$, where $\*Q$ be the set of all possible and feasible queries for the function. Our algorithm is described as follows \ref{alg:framework}. 

\begin{algorithm}[htb!]
\caption{Unified Framework}\label{alg:framework}
\KwData{$\*X$ \tcp*[r]{\footnotesize{Input}}}
\KwResult{$\*X_c$ \tcp*[r]{\footnotesize{Accurate Coreset}}}
\For{each $\*x \in \*X$}{
    Compute $\tilde{\*x} := \mathrm{vec}(g(\*x))$ \tcp*[r]{\footnotesize{$\mathrm{Kernelization}$}}
    $\tilde{\*X}^T = \begin{pmatrix} \tilde{\*X}^T, & \tilde{\*x} \end{pmatrix}$ 
}
$\{\*c,\*w\} := \mathrm{AccurateCoreset(\tilde{\*X})}$\;
Compute $\*X_c$ from $\*c,\*w$ and $\*X$ \tcp*[r]{\footnotesize{Identify selected samples}} 
\end{algorithm}

\paragraph{Algorithm Description:} For a given dataset and a function $f$, the above algorithm computes a new high dimensional structure $g(\*x)$ for every point $\*x \in \*X$. Here, the structure could be anything, e.g., a high dimensional vector (for $1$-mean), a matrix (for linear regression) or a tensor (for latent variable models) are just to name a few. Then it flattens the structure, i.e., $\tilde{\*x} = \mathrm{vec}(g(\*x))$ and appends it to a matrix $\tilde{\*X}$. This is called $\mathrm{Kernelization}$ whose property is defined in the following definition. 
\begin{definition}[$\mathrm{Kernelization}$]{\label{def:Kernelization}}
    The $\tilde{\*X}$ is $\mathrm{Kernelization}$ of $\*X$, such that for every $\*q \in \*Q$, there is a $\tilde{\*q} := \mathrm{vec}(g(\*q))$ that satisfies,
    \begin{equation}
        \sum_{\*x \in \*X}f(\*x,\*q) = \sum_{\tilde{\*x}} \tilde{\*x}^T\tilde{\*q} \label{eq:Kernelize}.
    \end{equation}
\end{definition}

For any machine learning algorithm, the $\mathrm{Kernelization}$ is the most challenging task in this framework. This essentially limits the construction of $\tilde{\*X}$ and, thereby, the existence of accurate coresets for the machine learning algorithm. 

In this paper, we present accurate coresets for both supervised and unsupervised machine learning models. We provide accurate coresets for $\ell_p$ regularized $\ell_p$ regression (supervised) for even valued $p$ and for generative models (unsupervised). An added regularization to any loss function reduces the complexity of the model. In the case of regularized regression, we show that the coreset size gradually decreases as the scale of the regularization parameter increases. This result subsumes ridge regression, which effectively improves the size of the accurate coreset. For generative models, we ensure lossless parameter estimation using a sublinear size accurate coreset. 

\subsection{Regularized Regression}
In this section, we present our result of accurate coreset for $\ell_p$ regularized $\ell_p$ regression for even valued $p$. Let $\{\*x_i\}_{i=1}^{n}$ be $n$ points in $\~R^d$ and $\*y \in \~R^n$ be the responses/labels of every point. The point set is represented by a matrix $\*X \in \~R^{n \times d}$. Let $\lambda > 0$ be a scalar (regularization parameter). The $\ell_p$ regularized $\ell_p$ regression problem optimizes the following loss function over every possible point $ \*w \in \~R^d$.
\begin{equation}
    \mathrm{RegLoss}(\*X,\*y,\lambda,\*w) := \|\*X\*w - \*y\|_p^p + \|\sqrt[p]{\lambda} \*I\*w\|_p^p \label{eq:RegLoss}
\end{equation}

We represent the usual regularization term $\lambda\|\*w\|_{p}^{p}$ as $\|\sqrt[p]{\lambda} \*I\*w\|_p^p$ for better relation  comparison with our coresets. 


A set $\{\*X_c,\*y_c\}$ is called an accurate coreset for the problem. $\*I_{c}$ is a square matrix such that the following guarantee for every $\*w \in \~R^d$,
\begin{equation}
     \mathrm{RegLoss}(\*X,\*y,\lambda,\*w) = \|\*X_c\*w - \*y_c\|_p^p + \|\sqrt[p]{\lambda}\*I_{c}\*x\|_{p}^{p}\label{eq:RegAccurate}.
\end{equation}

\subsubsection{Warm up: Linear Regression}
For simplicity, we start with simple linear regression, where $p = 2$, $\lambda = 0$, and assume $\*y = \*0$. The $\mathrm{RegLoss}(\*X,\*0,0,\*w) = \*1^T\tilde{\*X}\tilde{\*w}$, where
\begin{eqnarray}
    \tilde{\*X} = 
\begin{pmatrix}
\mathrm{vec}(\*x_1 \*x_1^T)^T \\
\mathrm{vec}(\*x_2 \*x_2^T)^T \\
\vdots \\
\mathrm{vec}(\*x_n \*x_n^T)^T
\end{pmatrix} \qquad \mbox{and} \qquad \tilde{\*w} = \mathrm{vec}(\*w \*w^T). \label{eq:RegressionKernel}
\end{eqnarray}
We $\mathrm{Kernelize}$ the points in $\*X$ and construct $\tilde{\*X}$. Thus, $\tilde{\*X} \in \~R^{n \times d^2}$ and similarly $\tilde{\*w} \in \~R^{d^2}$. Any algorithm that optimizes $\mathrm{RegLoss}(\*X,\*0,0,\*w)$, only searches over $\tilde{\*w} \in \mathrm{col}(\tilde{\*X})$, i.e., column space spanned by $\tilde{\*X}$. It is futile for any solution vector $\tilde{\*w}$ to have any component in $\~R^{d^2}$ that is not spanned $\mathrm{col}(\tilde{\*X})$. Being, orthogonal to $\mathrm{col}(\tilde{\*X})$, such components in $\*w$ will have no contribution in minimizing $\mathrm{RegLoss}(\*X,\*0,0,\*w)$. It is known that the size of the accurate coreset for linear regression is $O(d^2)$ \cite{maalouf2019fast}. This is because the size of the feasible (and useful) space for $\tilde{\*w}$ depends on $\mathrm{rank}(\tilde{\*X})$ which is $\binom{d+1}{2} = O(d^2)$.  

\subsubsection{Ridge Regression}
Now, for ridge regression where $\lambda > 0$, we have $\mathrm{RegLoss}(\*X,\*0,\lambda,\*w) = \*1^T\hat{\*X}\tilde{\*w}$. Here, the only update is in $\hat{\*X}$, where we append $d$ more rows to $\tilde{\*X}$ (see equation \eqref{eq:RegressionKernel}), as follows, 
\begin{equation}
    \hat{\*X} = 
\begin{pmatrix}
\tilde{\*X} \\
\lambda\cdot\mathrm{vec}(\*e_1 \*e_1^T)^T \\
\vdots \\
\lambda\cdot\mathrm{vec}(\*e_d \*e_d^T)^T
\end{pmatrix}. \label{eq:oldridgeKernel}
\end{equation}

So, the size of $\hat{\*X}$ is $(n+d) \times d^2$. Now, the convex hull of the point set also consists of $d$ points scaled by $\lambda$. Since the rank of $\hat{\*X}$ is still $\binom{d+1}{2}$, so by Theorem \ref{thm:caratheodory}, the required number of points in a subset that ensures the guarantees of accurate coreset is at most $\binom{d+1}{2}+1$, i.e., $\mathrm{rank}(\hat{\*X}) + 1$. It is well known that for any given matrix $\*X$, its rank is equal to the square of the Frobenius norm of its orthonormal column basis. Let $\hat{\*U}$ be the orthonormal column basis of $\hat{\*X}$. Since, the new rows in $\hat{\*X}$ could also be spanned by the rows in $\tilde{\*X}$ (i.e., $\mathrm{Kernelization}$ of $\*X$), so the $\mathrm{rank}(\tilde{\*X}) = \mathrm{rank}(\hat{\*X})$. Hence, $\|\hat{\*U}\|_F^2 = \binom{d+1}{2}$ and the set of points selected using Theorem \ref{thm:caratheodory} is at most $\binom{d+1}{2} + 1$ as established above. Now, based on the value of $\lambda$, some of the points from the last $d$ points (rows) may get selected in the final set. So, effectively, less than $\binom{d+1}{2}+1$ points are selected from the full dataset in the final set. We call these points the accurate coreset for the problem, and the weighted points selected from the last $d$ rows are used to appropriately redefine the regularization term. The actual number of samples that are selected from the full dataset is $\|\hat{\*U}_{(1\ldots n)}\|_F^2 + 1$, where the first $n$ rows of $\hat{\*U}$ is represented by $\hat{\*U}_{(1\ldots n)}$. 

Notice that the last $d$ rows of $\hat{\*X}$ are mutually independent and have equal norms. Since the $\mathrm{rank}(\tilde{\*X}) = \mathrm{rank}(\tilde{\*X}) = \binom{d+1}{2}$, hence the last $d$ rows of $\hat{\*X}$ only affects $d$ directions of $\tilde{\*X}$. Here, the main challenge is to analyze and quantify the part of the orthogonal column space of $\tilde{\*X}$ which is due to the first $n$ points, i.e., analyzing $\|\hat{\*U}_{(1\ldots n)}\|_F^2$. We carefully handle this with a simple and novel $\mathrm{Kernelization}$ technique, which we propose in the following lemma.
\begin{lemma}{\label{lem:RidgeRegression}}
    Let $\*X \in \~R^{n \times d}$ be $n$ points in $\~R^d$. Let $\lambda > 0$ be some regularization parameter. Let, $\tilde{\*X}$ be the $\mathrm{Kernelization}$ of $\*X$ as defined in equation \eqref{eq:RegressionKernel}. Let $\hat{\*X}^T = \begin{pmatrix} \tilde{\*X}^T, & \lambda \*T \end{pmatrix}$, where $\*T$ is a $d^2 \times d^2$ diagonal matrix. We define $T = \mathrm{diag}(\mathrm{vec}(\*M))$ where $\*M \in \{-1,+1\}^{d \times d}$ such that for every pair of indices $i \in [d]$ and $j \in [d]$, if $i < j$ then $M_{i,j} = -1$ else $M_{i,j} = 1$. Then for every $\*w \in \~R^d$ there is a $\tilde{\*w}$ as defined in equation \eqref{eq:RegressionKernel} we have,
    \begin{equation}
        \|\*X\*w\|_2^2 + \lambda\|\*w\|_2^2 = \*1^T\hat{\*X}\tilde{\*w} \label{eq:oldridgeKernelize}.
    \end{equation}
\end{lemma}

Before presenting the complete proof, we give a sketch of the proof, where we only discuss the $\mathrm{Kernelization}$ of the regularized term. The matrix $\*T$ is a $d^2 \times d^2$ diagonal matrix such that for any $\tilde{\*w}$ for its $\*w \in \~R^d$ we have $\*1^T\*T\tilde{\*w} = \|\*w\|_2^2$. Recall $\*w\*w^T$ is a $d \times d$ matrix. For every index in this matrix, we define the diagonal matrix $\*T$ from $\*M$. For every index in the lower triangle, i.e., below the diagonal term of $\*w\*w^T$ we assign a $-1$ to the corresponding index in $\*M$, and for the rest of the indices we assign a $+1$ in $\*M$. For, $\*T = \mathrm{diag}(\mathrm{vec}(\*M))$, the off-diagonal indices of $\*w\*w^T$ cancels out each other in $\*1^T\*T\tilde{\*w}$, and only the diagonal terms remain. Hence, we get $\*1^T\*T\tilde{\*w} = \|\*w\|_2^2$.
As an example, let $d = 2$ and $\*w^T = \begin{pmatrix} w_1, & w_2 \end{pmatrix}$. Then we define a matrix $\*T$ as follows,
\begin{equation}
    \hat{\*T} := \begin{pmatrix} 1 & 0 & 0 & 0 \\ 0 & -1 & 0 & 0 \\ 0 & 0 & 1 & 0 \\ 0 & 0 & 0 & 1 \end{pmatrix}. \nonumber 
\end{equation}
Since, both the off-diagonal terms in $\*w\*w^{T}$ are $(w_1\cdot w_2)$, which are the $2^{nd}$ and $3^{rd}$ indexed entries of $\tilde{\*w}$. So, with a $\*T$ defined in Lemma \ref{lem:RidgeRegression} we get $\*1^T\*T\tilde{\*w} = w_{1}^2+w_2^2 = \|\*w\|_2^2$.

\begin{proof}
    We first proof for the `unregularized regression' part $\|\*X\*w\|_2^2$. Notice that,
    \begin{eqnarray}
        \|\*X\*w\|_2^2 = \sum_{i=1}^{n}(\*x_i^T\*w)^2 = \sum_{i=1}^{n} \langle \*x_i\*x_i^T, \*w\*w^T \rangle \stackrel{(i)}{=} \sum_{i=1}^n   \mathrm{vec}(\*x_i\*x_i^T)^T \mathrm{vec}(\*w\*w^T) = \sum_{i=1}^n \tilde{\*x}_i^T \tilde{\*w} \stackrel{(ii)}{=} \|\tilde{\*X}\tilde{\*w}\|_2^2
    \end{eqnarray}

    The equality $(i)$ is by $\mathrm{Kernelization}$ of $\*X$. Now we prove the 'regularization' term. Notice, that
    \begin{eqnarray}
        \lambda \|\*w\|_2^2 = \lambda \sum_{i=1}^d w_i^2 \stackrel{(iii)}{=} \lambda \sum_{i=1}^d\sum_{j=1}^d M_{i,j}\cdot w_i\cdot w_j \nonumber
    \end{eqnarray}

    We get the equality $(iii)$ for appropriate values (as defined in the Lemma) of $M_{i,j}$ for every $i \in [d]$ and $j \in [d]$. Notice that for any fixed pair $\{i,j\}$ both $w_i\cdot w_j = w_j \cdot w_i$ by commutative rule. So, assigning $M_{i,j} = -M_{j,i}$ cancels out all the pairs, where $i \neq j$. Now, to consider the diagonal terms, we assign $M_{i,i} = 1$ for every $i \in [d]$. 
    
    Now, for the consistency in the operations of unregularized and regularized terms, we define $\*T$ a $\mathrm{Kernelization}$ of the regularized term. The matrix $\*T$ is a $d^2 \times d^2$ diagonal matrix. Here, $\*T = \mathrm{diag}(\mathrm{vec}(\*M))$. Such that the product $w_i \cdot w_j$ gets multiplied with $T_{k,k}$ where $k = (i-1)\cdot d + j$. Hence, for any $\tilde{\*w} = \mathrm{vec}(\*w\*w^T)$ for its $\*w \in \~R^d$ we have $\*1^T\*T\tilde{\*w} = \|\*w\|_2^2$. 

    Finally, by $\*1^T\*T\tilde{\*w} = \|\*w\|_2^2$ and the equation $(ii)$ we get the equation \eqref{eq:oldridgeKernelize}.
\end{proof}

The main advantage of this $\mathrm{Kernelization}$ is that $\*T$ being a diagonal matrix of size $d^2 \times d^2$ it universally affects all the directions in the column space of $\hat{\*X}$ with an equal measure. This phenomenon further enables us to accurately quantify the improved size of the accurate coreset for the ridge regression problem. 

We use the $\mathrm{Kernelization}$ defined in the Lemma \ref{lem:RidgeRegression} and formally state our algorithm for constructing accurate coresets for ridge regression. 
\begin{algorithm}[hbt!]
\caption{Coreset for Ridge Regression}\label{alg:ridge}
\SetAlgoLined
\KwData{$\*D = \{(\*x_i;y_i)\}_{i=1}^{n}, \lambda$\tcp*[r]{\footnotesize{$\*x_i \in \~R^d$; $y_i \in \~R$}}}
\KwResult{$\{\*c, \*w\}$ \tcp*[r]{\footnotesize{Accurate Coreset indices and their weights}}}
$\tilde{\*X} = \emptyset$, $\*M = \emptyset$\tcp*[r]{\footnotesize{Initialization}}
\For{$i = 1$ to $n$}{
    $\tilde{\*X} = \begin{pmatrix}
        \tilde{\*X} \\
        \mathrm{vec}(\*d_i\*d_i^T)^T 
    \end{pmatrix}$\tcp*[r]{\footnotesize{$\*d_i = \begin{pmatrix} \*x_i \\ y_i \end{pmatrix} \in \~R^{d+1}$}} 
}
\For{$i = 1$ to $d+1$}{
    \For{$j = 1$ to $d+1$}{
        \eIf{$i < j$}{
            $M_{i,j} =  -1$\tcp*[r]{\footnotesize{Handles lower triangular terms}}
            }
            {\eIf{$i == j == d+1$}{
            $M_{i,j} =  0$\tcp*[r]{\footnotesize{Handles last term}}
            }{
            $M_{i,j} =  1 $\tcp*[r]{\footnotesize{Handles remaining terms}}}
        } 
    }
}
$\*T = \mathrm{diag}(\mathrm{vec}(\*M))$ \tcp*[r]{\footnotesize{Diagonal matrix from $\*M$}}
$\hat{\*X}^T = \begin{pmatrix} \tilde{\*X}^T, & \lambda\*T \end{pmatrix}$ \tcp*[r]{\footnotesize{$\mathrm{Kernelization}$}}
$\{\*c,\*w\} := \mathrm{AccurateCoreset(\hat{\*X})}$\;
\end{algorithm}

\paragraph{Algorithm Description:} The algorithm \ref{alg:ridge} takes an input $\*X$, consisting of $n$ points in $\~R^d$ and its labels $\*y = \{y_1, \ldots, y_n\}$. For every $i \in [n]$, we augment the scalar $y_i$ at the bottom of $\*x_i$ and represent this augmentation as $\*d_i^T := \begin{pmatrix} \*x_i^T, & y_i \end{pmatrix}$. We denote these $n$ augmented vectors as a matrix $\*D \in \~R^{n \times (d+1)}$. The algorithm also takes the user-defined regularization parameter $\lambda$ as an input. First, it computes $\mathrm{Kernelization}$ of the rows of $\*D$ and stores them in $\tilde{\*X}$. For the matrix $\*D$ our queries are $\*q^T = \begin{pmatrix} \*w^T, & -1 \end{pmatrix}$ where $\*w \in \~R^d$. To ensure that $\*1^T\*T\tilde{\*q} = \|\*w\|_2^2$ the algorithm meticulously constructs $\*M$ for $\*T$, where $\tilde{\*q} = \mathrm{vec}(\*q\*q^T)$. For every term in the lower triangle of $\*q\*q^T$, it assigns a $-1$ to $\*t$, and for the rest (except for the bottom diagonal term) of terms, it assigns a $+1$ to $\*M$. Finally, for the bottom diagonal term of $\*q\*q^T$, it assigns a $0$ to $\*M$. Next, the algorithm constructs a diagonal matrix $\*T$ from $\*M$. Notice that the operation $\*1^T\*T\tilde{\*q} = \langle\*T,\*q\*q^T \rangle = \langle\mathrm{vec}(\*M),\tilde{\*q} \rangle$. Hence, the operation cancels out all the off-diagonal terms of $\*q\*q^T$ and does not consider them in the inner product. Further, due to $T_{d+1,d+1} = 0$ the inner product does not consider the bottom diagonal term of $\*q\*q^T$, i.e., $1$. As a result, we get $\*1^T\*T\tilde{\*q} = \|\*w\|_2^2$. Finally, the algorithm computes the final $\mathrm{Kerneliation}$ matrix $\hat{\*X}$ using $\tilde{\*X}, \*T$ and $\lambda$. Then, it calls the fast accurate coreset construction algorithm presented in \cite{maalouf2019fast}, which returns the indices $\*c$ of the selected points and their weights $\*w$. 

The indices of the coreset returned from the algorithm \ref{alg:ridge} ensure the following guarantee. 
\begin{theorem}{\label{thm:ridge}}
    Let, $\*D = \{(\*x_i;y_i)\}_{i=1}^{n}$ be the point/label pairs of $n$ samples. For every, $i \in [n]$ the point $\*x_i \in \~R^d$ and its corresponding label $y_i \in \~R$. Let $\lambda > 0$ be some positive scalar. Let $\{\sigma_1, \sigma_2, \ldots, \sigma_{\binom{d+2}{2}}\}$ be the singular values of $\tilde{\*X}$ defined in the algorithm \ref{alg:ridge}, then the algorithm computes the accurate coreset by selecting indices $\*c$ and their weights $\*w$, such that it only selects at most $O\left(\sum_{i=1}^{\binom{d+2}{2}} \frac{1}{1+\frac{\lambda^2}{\sigma_i^2}}\right)$ points from $\*D$ in time $O(nd^2 + d^8\log(n))$.     
\end{theorem}
\begin{proof}
    Consider the matrix $\hat{\*X}^{T} = \begin{pmatrix} \tilde{\*X}^{T}; \lambda \*T^{T} \end{pmatrix}$. It is important to note that the rank of $\tilde{\*X}$ is $\binom{d+2}{2}$. It can be easily verified from the fact that in our $\mathrm{Kernelization}$ of $\*D$ as $\tilde{\*X}$, for $\tilde{\*x}_i$ from every $\*d_i \in \*D$ only $\binom{d+2}{2}$ terms are unique, rest of them are equal or linear combinations of others. For example, if $d = 2$ such that for every $i \in [n]$ $\*d_i^T = \begin{pmatrix} x_{1,i}, & x_{2,i}, & y_i \end{pmatrix}$ then $\tilde{\*x}_i^T = \begin{pmatrix} x_{1,i}^2, & x_{1,i}x_{2,i}, & x_{1,i}y_{i}, & x_{2,i}x_{1,i}, & x_{2,i}^2, & x_{2,i}y_i, & y_ix_{1,i}, & y_ix_{2,i}, & y_i^2\end{pmatrix}$. Now in the matrix $\tilde{\*X}^T = \begin{pmatrix} \tilde{\*x}_1, & \tilde{\*x}_2, & \ldots, & \tilde{\*x}_n \end{pmatrix}$, notice that the second and fourth terms are exactly the same. Hence even though $\tilde{\*X} \in \~R^{n \times 9}$, i.e., a $n \times (d+1)^2$ size matrix, but its rank is $\binom{d+2}{2} = 6$. 

    Let $[\*U,\Sigma,\*V] = \mathrm{SVD}(\tilde{\*X})$. Here, we consider $(d+1)^2$ terns in our svd, i.e., $\*U \in \~R^{n \times (d+1)^2}$, $\Sigma \in \~R^{(d+1)^2 \times (d+1)^2}$ where only $\binom{d+2}{2}$ terms are non zero and rests are zeros and $\*V \in \~R^{(d+1)^2 \times (d+1)^2}$. Let $\*D$ is a diagonal matrix of size $(d+1)^2 \times (d+1)^2$ such that for every $i \in \left[\binom{d+2}{2}\right]$, $D_{i,i} = (\sigma_i^2 + \lambda^2)^{1/2}$ and for remaining $i \in [\binom{d+2}{2}+1, (d+1)^2 - 1]$ we set $D_{i,i} = \lambda^2$. We set $D_{(d+1)^2,(d+1)^2} = 0$. Now, we define a matrix $\hat{\*U} = \begin{pmatrix} \*U\Sigma\*D^{\dagger} \\ \lambda\hat{\*V}\*D^{\dagger} \end{pmatrix}$. Here for every $k \in [(d+1)^2]$, we set $\hat{\*V}_{k} = - \*V_k$ where $\hat{\*v}_k$ and $\*v_k$ are the $k^{th}$ column of $\hat{\*V}$ and $\*V$ respectively such that $\exists i \in [d+1]$, $j \in [i+1,d+1]$ and $k = (i-1)\cdot (d+1) + j$. Here $\*D^{\dagger}$ is the pseudo inverse of $\*D$. 
    
    Since, $\mathrm{rank}(\*T) = (d+1)^2-1 > \binom{d+2}{2}$, hence, the $\mathrm{rank}(\hat{\*X}) = (d+1)^2-1$. Notice, that $\hat{\*U}$ spans the column space of $\hat{\*X}$. This is because $\hat{\*U}\*D\*V^T = \hat{\*X}$. Now, we prove that $\hat{\*U}$ is the orthonormal column basis of $\hat{\*X}$. 
    \begin{eqnarray}
        \hat{\*U}^T\hat{\*U} = \*D^{\dagger}\Sigma \*U^T\*U \Sigma\*D^{\dagger} + \*D^{\dagger}\lambda^2 \hat{\*V}^T\hat{\*V}\*D^{\dagger} \stackrel{(i)}{=} \*D^{\dagger}\Sigma \Sigma\*D^{\dagger} + \*D^{\dagger}\lambda^2\*D^{\dagger} = \*D^{\dagger}\left(\Sigma^2 + \lambda^2\right)\*D^{\dagger} \stackrel{(ii)}{=} \*I_{(d+1)^2-1} \nonumber
    \end{eqnarray}
    In the equality $(i)$, we have $\*U^T\*U = \*I$ by the property of the orthonormal column basis. Further, in $\hat{\*V}$, we only changed the direction of a few vectors in $\*V$. However, this operation does not change the fact that every vector in $\*V$ is orthogonal to other vectors, and their norms are equal to $1$. Hence, we have $\hat{\*V}^{T}\*V = \*I$. Finally, the equality $(ii)$ by definition of $\*D$. 

    Now, we compute $\|\hat{\*U}_{(1, \ldots, n)}\|_F^2$ that measures the change in the model complexity, which reduces the effective rank of the data. 
    
    \begin{eqnarray}
        \|\hat{\*U}_{(1, \ldots, n)}\|_F^2 = \begin{Vmatrix} \*U\Sigma\*D^{-1} \end{Vmatrix}_F^2 \stackrel{(iii)}{=} \begin{Vmatrix} \Sigma\*D^{-1} \end{Vmatrix}_F^2 = \sum_{i=1}^{\binom{d+2}{2}} \frac{\sigma_i^2}{\sigma_i^2 + \lambda^2} = \sum_{i=1}^{\binom{d+2}{2}} \frac{1}{1 + \frac{\lambda^2}{\sigma_i^2}} \nonumber
    \end{eqnarray}

    In the equality $(iii)$, we use the property of orthonormal column basis $\*U$ of $\tilde{\*X}$. Due, to $\lambda > 0$, as the rank of the data $\*D$ in the space $\~R^{(d+1)^2}$ reduces from $\binom{d+2}{2}$ to $\sum_{i=1}^{\binom{d+2}{2}} \frac{1}{1 + \frac{\lambda^2}{\sigma_i^2}}$, hence, by the theorem \ref{thm:caratheodory} our algorithm selects at most $\sum_{i=1}^{\binom{d+2}{2}} \frac{1}{1 + \frac{\lambda^2}{\sigma_i^2}} + 1$ points from $\*D$. 

    The running time of our coreset construction Algorithm \ref{alg:fast_caratheodory} is $O\left(nd^2 + d^8 \log(n)\right)$ due to the fast Caratheodory algorithm from \cite{maalouf2019fast}. 
\end{proof}

It is important to note that in the above theorem, the analysis of the size of the accurate coreset was possible due to the novel $\mathrm{Kernelization}$ of the regularization term (Lemma \ref{lem:ReguRegression}). It ensures a universal effect on all the directions that are spanned by $\tilde{\*X}$, 

In practice, the returned indices $\*c$ and the weight function $\*w$ are used to construct both the accurate coreset and the $\*I_{c}$ matrix, thereby getting a lossless trained model from them. We discuss this construction in detail.
\subsubsubsection{Coreset Construction:} Here, we describe the coreset construction using the indices retired from the algorithm \ref{alg:ridge}. Recall that $\*c$ and $\*w$ are the set of selected indices and their associated weights. Let $i$ be any selected index in $\*c$, if $i \in [n]$ then due to one-to-one mapping between the original points $\*x$ and its $\mathrm{Kernelization}$ point $\tilde{\*x}$ we select point $\*x_i$ from $\*X$ and $y_i$ from $\*y$. The selected point and label pair $(\*x_{i}, y_{i})$ are scaled by $\sqrt{\*w(i)}$. We store these weighted pairs in $\*X_c$ and $\*y_c$. Next, if $i \in [n+1, n+d^2]$, then define $\*M_i \in \~R^{d \times d}$ such that for the corresponding index, we set the same values as in $\*M$ (as defined in the algorithm \ref{alg:ridge}) and for rest of the indices we set it to $0$. Scale this matrix by $\*w(i)$. We represent them as $\lambda\*I_c$. Finally, we solve the ridge regression problem as follows,
\begin{equation*}
    \*x^* := \left(\*X_c^T\*X_c + \lambda\*I_c\right)^{-1}\*X_c^T\*y_c.
\end{equation*}

\subsubsection{$\ell_p$ Regularized $\ell_p$ Regression}
In this section, we generalize our result of ridge regression for any even valued $p$, i.e., accurate coreset for $\ell_p$ regularized $\ell_p$ regression. Let $\*X \in \~R^{n \times d}$ be the dataset and $\*y \in \~R^n$ be its label. Let $\lambda > 0$ be some regularization parameter. Now, for a fixed even valued $p$ the regularized regression minimizes $\mathrm{RegLoss}(\*X,\*y,\lambda,\*w)$ over all possible $\*w \in \~R^d$. 

Let  $\*D = \{(\*x_i,y_i)\}_{i=1}^{n}$ be the dataset of $n$ samples. Let $\tilde{\*X}^T = \begin{pmatrix} \mathrm{vec}(\*d_1 \otimes^p), & \mathrm{vec}(\*d_2 \otimes^p), & \cdots, & \mathrm{vec}(\*d_n \otimes^p)\end{pmatrix}$ where $\*d_i^T = \begin{pmatrix} \*x_i^{T}, & y_i\end{pmatrix}$.

Next, we design a new $\mathrm{Kernelization}$ such that it retains $\mathrm{RegLoss}(\*X,\*y,\lambda,\*w)$ for every $\*w \in \~R^d$. The $\mathrm{Kernelization}$ technique is stated in the following lemma. 
\begin{lemma}{\label{lem:ReguRegression}}
    Given a dataset $\*D = \{(\*x_i,y_i)\}_{i=1}^{n}$ and $\lambda > 0$, $\hat{\*X} \in \~R^{n \times (d+1)^p}$ be the $\mathrm{Kernelization}$ of 
$\ell_p$ regularized $\ell_p$ regression. It is defined as, $\hat{\*X}^T := \begin{pmatrix} \tilde{\*X}^T, & \*T\end{pmatrix}$, where, $\tilde{\*X}$ be the $\mathrm{Kerneliation}$ of $\*D$ as defined above. Let $T = \mathrm{diag}(\mathrm{vec}(\@M))$ be a $d^p \times d^p$ diagonal matrix  where $\@M \in \{-1,+1\}^{d \times d \times \cdot \times d}$ a $p$-order tensor. For every set of $p$ indices $\{i_1, i_2, \ldots, i_p\} \in [d+1]$, let $ind$ be the all possible unique combinations of index sets. Let $ind = ind1 \cup ind2$ such that $ind1 \cap ind2 = \emptyset$ and $|ind1| = |ind2|$. When all the indices in the set are not equal, set $M_{j} = -1$ where $j \in ind1$ and for $j \in ind2$ set $M_{j} = 1$. For every $i \in [d]$ set, $M_{i,\ldots,i} = 1$. Then for every $\*w \in \~R^d$ there is a $\tilde{\*w} := \mathrm{vec}(\begin{pmatrix} \*w^T, & -1\end{pmatrix} \otimes^p)$ we have,
    \begin{equation}
        \|\*X\*w - \*y\|_p^p + \lambda\|\*w\|_p^p = \*1^{T}\hat{\*X}\tilde{\*w} \label{eq:lpRegugeKernelize}
    \end{equation}
\end{lemma}

\begin{proof}
    This proof is similar to the proof of Lemma \ref{lem:RidgeRegression}. First we proof for the `unregularized regression' part $\|\*X\*w - \*y\|_p^p = \|\*D\*z\|_p^p$, where $\*D = \begin{pmatrix} \*X, & \*y \end{pmatrix}$ and $\*z = \begin{pmatrix} \*w \\ -1 \end{pmatrix}$. Notice that,
    \begin{eqnarray}
        \|\*D\*z\|_p^p \stackrel{(i)}{=} \sum_{i=1}^{n}(\*d_i^T\*z)^p = \sum_{i=1}^{n} \langle \*d_i \otimes^p, \*z\otimes^p \rangle \stackrel{(ii)}{=} \sum_{i=1}^n   \mathrm{vec}(\*d_i \otimes^p)^T \mathrm{vec}(\*z \otimes^p) = \sum_{i=1}^n \tilde{\*x}_i^T \tilde{\*z} \stackrel{(iii)}{=} \|\tilde{\*X}\tilde{\*z}\|_p^p \nonumber
    \end{eqnarray}

    In equality $(i)$ we consider $\*d_i^T$ as the $i^{th}$ row of $\*D$ for $1 \leq i \leq n$. The equality $(ii)$ is by $\mathrm{Kernelization}$ of $\*D$. Now we prove the 'regularization' term. Notice, that
    \begin{eqnarray}
        \lambda \|\*w\|_p^p = \lambda \sum_{i=1}^d w_i^p \stackrel{(iv)}{=} \lambda \sum_{i_1=1}^{d+1}\sum_{i_2=1}^{d+1}\cdots\sum_{i_p=1}^{d+1} M_{i_1, i_2, \ldots,i_p}\cdot z_{i_1} \cdot z_{i_2} \cdots z_{i_p} \nonumber
    \end{eqnarray}

    We get the equality $(iv)$ for appropriate values (as defined in the Lemma) in $\@M$. It is important to note that for any tuple $\{i_1,i_2,\ldots,i_p\}$ the number of unique ordered combinations is even, unless $i_1 = i_2 = \cdots = i_p$. Hence, we set half of them to $-1$ and the rest half to $+1$ in $\@M$. This will effectively cancel out the off-diagonal terms and retain only the diagonal terms in our operation. Further, to ensure that $z_{d+1}^p$ is not been contributed in our operation, we force $M_{d+1, \ldots, d+1} = 0$. 
    
    Now, for the consistency in the operations of unregularized and regularized terms, we define $\*T$ a $\mathrm{Kernelization}$ of the regularized term. The matrix $\*T$ is a $d^p \times d^p$ diagonal matrix. Here, $\*T = \mathrm{diag}(\mathrm{vec}(\@M))$. Such that the product $z_{i_1} \cdot z_{i_2}\cdots z_{i_p}$ gets multiplied with $T_{k,k}$ where $k = (i_1-1)\cdot d^{p-1} + (i_2-1)\cdot d^{p-2} + \cdots + (i_{p-2} - 1)\cdot d^2 + (i_{p-1}-1)\cdot d + i_p$. Hence, for any $\tilde{\*z} = \mathrm{vec}(\*z\otimes^p)$ for its $\*w \in \~R^d$ we have $\*1^T\*T\tilde{\*z} = \|\*w\|_p^p$. 

    Finally, by $\*1^T\*T\tilde{\*w} = \|\*w\|_2^2$ and the equation $(ii)$ we get the equation \eqref{eq:lpRegugeKernelize}.
\end{proof}

Notice, that $\hat{\*X}$ in the algorithm will have rank at most $\binom{d+p-1}{p}$. Now, using the previous lemma, we state our accurate coreset construction algorithm for $\ell_p$ regularized $\ell_p$ regression for a general even valued $p$. 
\begin{algorithm}[hbt!]
\caption{Coreset for Regularized Regression}\label{alg:lpReguRegression}
\SetKwProg{Fn}{Compute}{ such that}{end}
\KwData{$\*D = \{(\*x_i;y_i)\}_{i=1}^{n}, \lambda, p$\tcp*[r]{\footnotesize{$\*x_i \in \~R^d$; $y_i \in \~R$}}}
\KwResult{$\{\*c,\*w\}$ \tcp*[r]{\footnotesize{Accurate Coreset indices and their weights}}}
$\tilde{\*X} = \emptyset$, $\*M = \emptyset$\tcp*[r]{\footnotesize{Initialization}}
\For{$i = 1$ to $n$}{
    $\tilde{\*X} = \begin{pmatrix}
        \tilde{\*X} \\
        \mathrm{vec}(\*d_i \otimes^p)^T 
    \end{pmatrix}$\tcp*[r]{\footnotesize{$\*d_i = \begin{pmatrix} \*x_i \\ y_i \end{pmatrix} \in \~R^{d+1}$}} 
}
\For{every tuple $t = \{i_1, i_2, \ldots, i_p\} \in [d+1]$}{
    $ind := \mbox{Combinations}(t)$ \tcp*[r]{\footnotesize{set of all unique combinations of the indices in $t$}}
    \Fn{$ind1$ and $ind2$}{
        $ind1 \cup ind2 == ind$\;
        $ind1 \cap ind2 == \emptyset$\;
        $|ind1| == |ind2|$\;
    }
    \For{every $j \in ind$}{
        \eIf{$j \in ind1$}{
            Set $M_{j} = -1$ \tcp*[r]{\footnotesize{-1 for half of the combinations}}
            }{
                Set $M_{j} = 1$ \tcp*[r]{\footnotesize{1 for the rest}}
            }
        }
        \If{$j == \{i,i,\ldots, i\}$}{
            Set $M_{j} = 1$ \tcp*[r]{\footnotesize{1 for the diagonals}}
        }
        \If{$j == \{d+1,d+1,\ldots, d+1\}$}{
            Set $M_{j} = 0$
        }
    }
$\*T = \mathrm{diag}(\mathrm{vec}(\@M))$ \tcp*[r]{\footnotesize{Diagonal matrix from tensor $\@M$}}
$\hat{\*X}^T = \begin{pmatrix} \tilde{\*X}^T, & \lambda\*T \end{pmatrix}$ \tcp*[r]{\footnotesize{$\mathrm{Kernelization}$}}
$\{\*c, \*w\} := \mathrm{AccurateCoreset(\hat{\*X})}$\; 
\end{algorithm}

The indices of the coreset returned from the algorithm \ref{alg:ridge} ensures the following guarantee. 

\begin{theorem}{\label{thm:lpReguRegression}}
    Let, $\*D = \{(\*x_i;y_i)\}_{i=1}^{n}$ be the point/label pairs of $n$ samples. For every, $i \in [n]$ the point $\*x_i$ is in $\~R^d$ and its corresponding label $y_i \in \~R$. Let $\lambda > 0$ be some positive scalar. Let $\{\sigma_1, \sigma_2, \ldots, \sigma_{\binom{d+p-1}{p}}\}$ be the singular values of $\tilde{\*X}$ defined in the algorithm \ref{alg:lpReguRegression}, then the algorithm computes the accurate coreset by selecting indices $\*c$ and their weights $\*w$, such that it only selects at m
    ost $O\left(\sum_{i=1}^{\binom{d+p-1}{p}} \frac{1}{1+\frac{\lambda^2}{\sigma_i^2}}\right)$ points from $\*D$ in time $O(nd^p + d^{4p}\log(n))$.     
\end{theorem}

\begin{proof}
    Consider the matrix $\hat{\*X} = \begin{pmatrix} \tilde{\*X} \\ \lambda \*T \end{pmatrix}$. It is important to note that the rank of $\tilde{\*X}$ is $\binom{d+p}{p}$. It can be easily verified from the fact that in our $\mathrm{Kernelization}$ of $\*D$ as $\tilde{\*X}$, for $\tilde{\*d}_i = \mathrm{vec}(\*d_i \otimes^p)$ from every $\*d_i \in \*D$ only $\binom{d+p}{p}$ terms are unique, rest of them are equal or linear combinations of others. This is because for any tuple $\{i_1,i_2, \ldots, i_p\}$ each in $[d+1]$, although its permutation represents a different entry in $\tilde{\*x}_i$, their values are the same (i.e., $d_{i,i_1}\cdot d_{i,i_2} \cdots d_{i,i_p}$) due to the commutative law of multiplication. For every $i \in [n]$ the vector $d_i$ is defined as $\*d_i^T = \begin{pmatrix} x_{1,i}, & x_{2,i}, & \cdots, & x_{d,i}, & y_i \end{pmatrix}$ then $\tilde{\*x}_i^T = \begin{pmatrix} x_{1,i}^p, & x_{1,i}^{p-1}\cdot x_{2,i}, &  \cdots, & y_{i}^p\end{pmatrix}$. Now in the matrix $\tilde{\*X}^T = \begin{pmatrix} \tilde{\*x}_1, & \tilde{\*x}_2, & \cdots, & \tilde{\*x}_n \end{pmatrix}$. Notice that not all the values are unique. Hence even though $\tilde{\*X} \in \~R^{n \times (d+1)^p}$, but its rank is $\binom{d+p}{p}$. 
    
    Let $[\*U,\Sigma,\*V] = \mathrm{SVD}(\tilde{\*X})$. Here, we consider $(d+1)^p$ terns in our svd, i.e., $\*U \in \~R^{n \times (d+1)^p}$, $\Sigma \in \~R^{(d+1)^p \times (d+1)^p}$ where only $\binom{d+p}{p}$ terms are non zero and rests are zeros and $\*V \in \~R^{(d+1)^p \times (d+1)^p}$. Let $\*D$ is a diagonal matrix of size $(d+1)^p \times (d+1)^p$ such that for every $i \in \left[\binom{d+p}{p}\right]$, $D_{i,i} = (\sigma_i^2 + \lambda^2)^{1/2}$ and for every $i \in [\binom{d+p}{p}+1, (d+1)^p - 1]$ we set $D_{i,i} = \lambda^2$. We set, $D_{(d+1)^p,(d+1)^p} = 0$. Now, we define a matrix $\hat{\*U} = \begin{pmatrix} \*U\Sigma\*D^{\dagger} \\ \lambda\hat{\*V}\*D^{\dagger} \end{pmatrix}$. For a tuple $\{i_1,i_2, \ldots, i_p\}$, notice that there are always even numbers of permutations. So we assign half of the permutations as set $ind1$ and the rest half as $ind2$. Now, for every $k \in ind1$, we set $\hat{\*V}_{k} = - \*V_k$ and for the every $k \in ind2$, we set $\hat{\*V}_{k} = \*V_k$ where $\hat{\*v}_k$ and $\*v_k$ are the $k^{th}$ column of $\hat{\*V}$ and $\*V$ respectively
    
    Since, $\mathrm{rank}(\*T) = (d+1)^p - 1 > \binom{d+p}{p}$, hence, the $\mathrm{rank}(\hat{\*X}) = (d+1)^p - 1$. Notice, that $\hat{\*U}$ spans the column space of $\hat{\*X}$. This is because $\hat{\*U}\*D\*V^T = \hat{\*X}$. Now, we prove that $\hat{\*U}$ is the orthonormal column basis of $\hat{\*X}$.
    \begin{eqnarray}
        \hat{\*U}^T\hat{\*U} = \*D^{\dagger}\Sigma \*U^T\*U \Sigma\*D^{\dagger} + \*D^{\dagger}\lambda^2 \hat{\*V}^T\hat{\*V}\*D^{\dagger} \stackrel{(i)}{=} \*D^{\dagger}\Sigma \Sigma\*D^{\dagger} + \*D^{\dagger}\lambda^2\*D^{\dagger} = \*D^{\dagger}\left(\Sigma^2 + \lambda^2\right)\*D^{\dagger} \stackrel{(ii)}{=} \*I_{(d+1)^p-1} \nonumber
    \end{eqnarray}
    In the equality $(i)$, we have $\*U^T\*U = \*I$ by the property of the orthonormal column basis. Further, in $\hat{\*V}$, we only changed the direction of a few vectors in $\*V$. However, this operation does not change the fact that every vector in $\*V$ is orthogonal to other vectors, and their norms are equal to $1$. Hence, we have $\hat{\*V}^{T}\*V = \*I$. Finally, the equality $(ii)$ by definition of $\*D$. 

    Now, we compute $\|\hat{\*U}_{(1, \ldots, n)}\|_F^2$ that measures the change in the model complexity, which reduces the effective rank of the data. 
    
    \begin{eqnarray}
        \|\hat{\*U}_{(1, \ldots, n)}\|_F^2 = \begin{Vmatrix} \*U\Sigma\*D^{-1} \end{Vmatrix}_F^2 \stackrel{(iii)}{=} \begin{Vmatrix} \Sigma\*D^{-1} \end{Vmatrix}_F^2 = \sum_{i=1}^{\binom{d+1}{2}} \frac{\sigma_i^2}{\sigma_i^2 + \lambda^2} = \sum_{i=1}^{\binom{d+1}{2}} \frac{1}{1 + \frac{\lambda^2}{\sigma_i^2}} \nonumber
    \end{eqnarray}

    In the equality $(iii)$, we use the property of orthonormal column basis $\*U$ of $\tilde{\*X}$. Due, to $\lambda > 0$, as the rank of the data $\*D$ in the space $\~R^{(d+1)^p}$ reduces from $\binom{d+p}{p}$ to $\sum_{i=1}^{\binom{d+p}{p}} \frac{1}{1 + \frac{\lambda^2}{\sigma_i^2}}$, hence, by the theorem \ref{thm:caratheodory} our algorithm selects at most $\sum_{i=1}^{\binom{d+p}{p}} \frac{1}{1 + \frac{\lambda^2}{\sigma_i^2}} + 1$ points from $\*D$. 

    The running time of our coreset construction algorithm is $O\left(nd^p + d^{4p} \log(n)\right)$ due to the fast Caratheodory algorithm \ref{alg:fast_caratheodory} from \cite{maalouf2019fast}. 
\end{proof}

\subsection{Generative Models}
The latent variable model is a generative model where the goal is to realize the unknown parameters (or latent variables) that are believed to be used to generate the observed dataset. The common latent variable models include problems such as Gaussian Mixture Models (GMM), Hidden Markov Models (HMM), Single Topic Modeling, Latent Dirichlet Allocations (LDA), etc. 
It is assumed that every sampled point is independent or conditionally independent and that they are identically distributed over the latent variable space. \cite{hsu2013learning} constructed higher order moments using the datasets such that the over expectation, the moments are equal to the higher order moments of the latent variables. 

For completeness, we describe one of the simple latent variable model problems called single-topic modeling. Every document is a bag of words representing only one topic. A topic is a unique distribution of the words in the vocabulary space. It is considered that during a document generation, first, a topic is sampled with some probability (which identifies the number of documents from each topic), and then a set of words are sampled from the vocabulary space with the distribution that identifies the selected topic. 

Let there are $n$ documents represented by $\@D = \{\*D_1, \ldots, \*D_n\}$ such that every $i^{th}$ document has $n_i$ words $\{\*w_1,\ldots, \*w_{n_i}\} \in \~R^{d}$ where $d$ is the size of vocabulary. So, $\*D_i \in \~R^{n_i \times d}$ size matrix. Now, in the same document, if the $q^{th}$ word is the $t^{th}$ word in the vocabulary, then the $q^{th}$ row of $\*D_i$ is $\*w_q = \*e_t$, where $\*e_t$ is the $t^{th}$ vector in the standard basis of $\*I_d$. Let there are $k$ topics, where $\mu_j \in \Delta^{d-1}$ is the distribution over the words in the vocabulary space for topic $j \in k$. Then, $\~E[\*x_q| h=j] = \mu_j$. Further, $\~E[\*x_q|h=j] \~E[\*x_r|h=j]^T= \mu \mu^T$. If the distribution of documents in $\@D$ follows a distribution over topics $\gamma \in \Delta^{k-1}$ then,
\begin{eqnarray}
    &&\*t_1 := \~E_{\*w_1}[\*w_1] = \sum_{j=1}^{k} \gamma_j\cdot \mu_j  \nonumber \\
    &&\*T_2 := \~E_{\*w_1,\*w_2}[\*w_1\*w_2^T] = \sum_{j=1}^{k} \gamma_j\cdot \mu_j \mu_j^{T} \nonumber \\
    &&\@T_p := \~E_{\*w_1,\*w_2,\ldots \*w_p}[\*w_1\otimes\ldots\otimes\*w_p] = \sum_{j=1}^{k} \gamma_j\cdot \mu_j \otimes^p \nonumber
\end{eqnarray}

In noisy data, a low-rank decomposition of the above terms is enough to approximate the latent variables $\{\gamma_i,\mu_i\}_{i=1}^{k}$. Now, a low-rank decomposition is not unique for a sum of vectors (first-moment) or a sum of matrices (second-moment). However, it is unique for higher-order moments under a mild assumption that no two decomposition components are linearly dependent on each other \cite{kruskal1977three}. So, when $k < d$, it is enough to work with $\*T_2$ and $\@T_3$. Similar to matrix decomposition, a simple power iteration method can be used for tensor decomposition. However, for power iteration to work out, it is important to ensure that the components of the decomposition are orthogonal to each other. For this we transform $\tilde{\@T}_3 = \@T_3(\*M,\*M,\*M)$, where $\*M^T\*T_2\*M = \*I_k$. The transformation is called $\mathrm{Whitening}$. It reduces the size of the tensor $\@T_3$ from $d \times d \times d$ to a tensor $\tilde{\@T}_3$ of size $k \times k \times k$. This also ensures that the components of tensor decomposition of $\tilde{\@T}_3$ are orthogonal to each other. 

For other latent variable models such as GMM, HMM, LDA etc, the higher order moments are computed very carefully \cite{anandkumar2014tensor}. For all of them, during tensor decomposition using the power iteration method, {\em tensor contraction} is one of the most important operations in the tensor decomposition using the power iteration method. It starts with a random vector $\*x$ in $\~R^{k}$ and update it as follows,
\begin{equation*}
    \*x := \frac{\@T_3(\*x,\*x, \*I_k)}{\|\@T_3(\*x,\*x, \*I_k)\|_2}.
\end{equation*}

\subsubsection{Coreset for Tensor Contraction}
So, having a coreset for tensor contraction is sufficient for approximating the latent variables of the generative models. Here, we present accurate coresets for this problem. Coresets for tensor contraction were introduced in \cite{chhaya2020streaming}. However, the coreset guarantees are relative only when $p$ is an even-valued integer, and for the odd values of $p$, their guarantees were additive error approximations. So, the difference between the tensor contraction on the original data and on coresets can suffer significantly. 

In the following lemma, we state our $\mathrm{Kernelization}$ for the tensor contraction problem.  
\begin{lemma}{\label{lem:lvm}}
    Let $\{\*x_1, \*x_2, \ldots, \*x_n\}$ be $n$ points in $\~R^d$ represented by a matrix $\*X \in \~R^{n \times d}$. For any $p \in \~Z_+$, let $\@T_p = \sum_{i=1}^n \*x_i \otimes^p$. Let, $\hat{\*X}$ be the $\mathrm{Kernelization}$ of $\*X$ defined as, $\hat{\*X}^T = \begin{pmatrix} \mathrm{vec}(\*x_1\otimes^p), & \ldots & \mathrm{vec}(\*x_n\otimes^p) \end{pmatrix}$. Then for every $\*x \in \~R^d$ there is a $\tilde{\*x} := \mathrm{vec}(\*x\otimes^p)^T$ such that,
    \begin{equation}
        \@T_p(\*x, \ldots, \*x) =  \*1^T\hat{\*X}\tilde{\*x}\label{eq:lvmKernelize}.
    \end{equation}
\end{lemma}

\begin{proof}
    This proof is similar to the proof of Lemma \ref{lem:ReguRegression}. Let $\*D = \{\*x_1, \*x_2, \ldots, \*x_n\}$. Notice that for any $\*z \in \~R^d$ the tensor contraction on $\@T_3$ due to $\*z$ is, 
    \begin{eqnarray}
        \@T_3(\*z,\*z, \ldots, \*z) \stackrel{(i)}{=} \sum_{i=1}^{n} (\*x_i^T\*z)^p = \sum_{i=1}^{n} \langle \*x_i \otimes^p, \*z\otimes^p \rangle \stackrel{(ii)}{=} \sum_{i=1}^n  \mathrm{vec}(\*x_i \otimes^p)^T \mathrm{vec}(\*z \otimes^p) = \sum_{i=1}^n \tilde{\*x}_i^T \tilde{\*z} \stackrel{(iii)}{=} \|\tilde{\*X}\tilde{\*z}\|_p^p \nonumber
    \end{eqnarray}

    In equality $(i)$ we consider $\*x_i^T$ as the $i^{th}$ data point or $i^{th}$ row of $\*D$ for $1 \leq i \leq n$. The equality $(ii)$ is by $\mathrm{Kernelization}$ of $\*D$.
\end{proof}

Using the above lemma, we present our accurate coreset construction algorithm for the latent variable model problem.

\begin{algorithm}[hbt!]
\caption{Coreset for Latent Variable Models}\label{alg:lvm}
\KwData{$\*X = \{\*x_1, \ldots, \*x_n\}, k < d$\;} 
\KwResult{$\{\*X_c, \*w\}$ \tcp*[r]{\footnotesize{Accurate Coreset}}}
$\hat{\*X} := \emptyset$\;
Compute $\*T_2$ \tcp*[r]{\footnotesize{Empirical average of second order moment}} 
$[\*U,\Sigma,\*V] := \mathrm{SVD}(\*T_2,k)$ \tcp*[r]{\footnotesize{Top $k$ singular value/vector pairs}} 
$\*M := \*V^{T}\sqrt{\Sigma}$\;
\For{$i = 1$ to $n$}{
    Compute $\@T$ for $\*x_i$ \tcp*[r]{\footnotesize{Third order moment}}
    $\tilde{\@T} = \frac{1}{n}\@T(\*M,\*M,\*M)$ \tcp*[r]{\footnotesize{$\mathrm{Whitening}$}} 
    $\hat{\*X}^T = \begin{pmatrix} \hat{\*X}^T, & \mathrm{vec}(\tilde{\@T}) \end{pmatrix}$ 
}
$\{\*c,\*w\} := \mathrm{AccurateCoreset(\hat{\*X})}$\;
Compute $\*X_c$ from $\*c,\*w$ and $\*X$ \tcp*[r]{\footnotesize{Identify selected samples}} 
\end{algorithm}


The above algorithm returns a coreset of size $O(k^3)$. Since it is an accurate coreset with a sublinear number of samples, one can learn exactly the same parameters that could have been learned from the complete dataset. The coreset returned from Algorithm \ref{alg:lvm} ensures the following guarantees. 
\begin{theorem}{\label{thm:lvm}}
    Let, $\*X = \{(\*x_i)\}_{i=1}^{n}$ be the $n$ samples for a latent variable model problem. For $k < d$, the algorithm \ref{alg:lvm}, computes the accurate coreset $\*X_c$ and their weights $\*w$, such that it only selects at most $\binom{k+2}{3} + 1$ samples from $\*X$ in time $O(nk^3 + k^{12}\log(n))$.
\end{theorem}

\begin{proof}
    The proof is simplified version of the proof of Theorem \ref{thm:lpReguRegression}, where $p=3$, $\lambda = 0$ and $\*y = \*0$. 
    
    Let $\*X$ be the dataset and $\hat{\*X}$ be the matrix as defined in Algorithm \ref{alg:lvm} for $p=3$. Let $\*M$ be the whitening matrix. It is important to note that the rank of $\hat{\*X}$ is $\binom{k+2}{3}$. It can be easily verified from the fact that in our $\mathrm{Kernelization}$ of $\tilde{\*X}$ as $\hat{\*X}$ where $\tilde{\*X} = \*X\*M$. Here, for $\hat{\*x}_i = \mathrm{vec}(\tilde{\*x}_i \otimes^3) = \mathrm{vec}(\*x_i^T\*M \otimes^3)$ from every $\*x_i \in \*X$ only $\binom{k+2}{3}$ terms are unique, rest of them are equal or linear combinations of others. This is because of the same argument as there in the proof of Theorem \ref{thm:lpReguRegression}, i.e., any tuple $\{i_1,i_2, i_3\}$ each in $[k]$, although its permutation represents a different entry in $\hat{\*x}_i$, their values are the same (i.e., $\tilde{x}_{i,i_1}\cdot \tilde{x}_{i,i_2} \cdot \tilde{x}_{i,i_3}$) due to the commutative law of multiplication. For every $i \in [n]$ $\tilde{\*x}_i = \begin{pmatrix} \tilde{x}_{1,i} \\ \tilde{x}_{2,i} \\ \vdots \\ \tilde{x}_{d,i}\end{pmatrix}$ then $\hat{\*x}_i = \begin{pmatrix} \tilde{x}_{1,i}^3 \\ \tilde{x}_{1,i}^{2}\cdot \tilde{x}_{2,i} \\ \vdots \\ \tilde{x}_{d,i}^3\end{pmatrix}$. Now in the matrix $\hat{\*X} = \begin{pmatrix} \hat{\*x}_1^T \\ \hat{\*x}_2^T \\ \vdots \\ \hat{\*x}_n^T \end{pmatrix}$. Even though $\hat{\*X} \in \~R^{n \times k^3}$, but its rank is $\binom{k+2}{3}$. So, by the theorem \ref{thm:caratheodory} our algorithm selects at most $\binom{k+2}{3} + 1$ points from $\*X$. 

    The running time of our coreset construction algorithm is $O\left(nk^3 + k^{12} \log(n)\right)$ due to the fast Caratheodory algorithm \ref{alg:fast_caratheodory} from \cite{maalouf2019fast}. 
\end{proof}

For the case of single topic modeling, the algorithm \ref{alg:lvm} would return accurate coresets that only select at most $\binom{k+2}{3} + 1$ documents from $\@D$. 

From the lemma \ref{lem:lvm} and Theorem \ref{thm:lvm}, the following corollary generalizes the size of accurate coreset for tensor contraction of any $p$-order tensor.
\begin{corollary}{\label{cor:TensorContraction}}
    For $p$-order tensor contraction on a tensor of size $d \times d \times \ldots \times d$ as shown in Lemma \ref{lem:lvm}, the size of an accurate coreset for this problem is at most $\binom{d+p-1}{p} + 1$.
\end{corollary}

\begin{proof}
    From lemma \ref{lem:lvm}, as the rank of $\mathrm{Kernelization}$ of $p$-order tensor is $\binom{d+p+1}{p}$, so, by the theorem \ref{thm:caratheodory} our algorithm selects at most $\binom{d+p-1}{p} + 1$ points from the data set. 
\end{proof}

%% file: section/experiment.tex
\section{Experiments}{\label{sec:exp}}
In this section, we support our theoretical claims with empirical evaluations on real datasets. We checked the performance of our algorithms for ridge regression and GMM. Here, we have used various real datasets.

\subsection{Ridge Regression}
\begin{figure*}[h]
\vspace{.3in}
\centerline{\fbox{\includegraphics[width=0.25\linewidth]{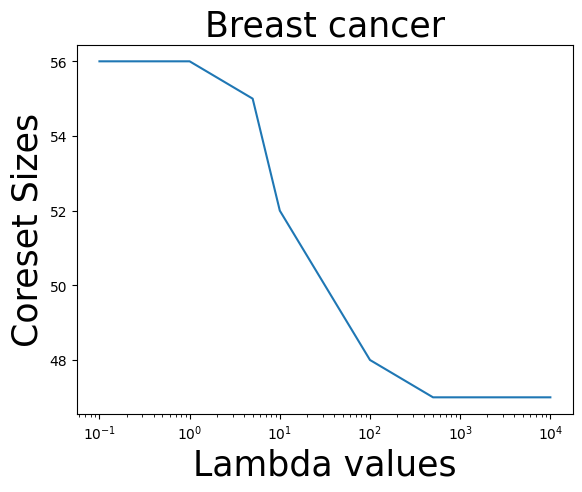}\includegraphics[width=0.25\linewidth]{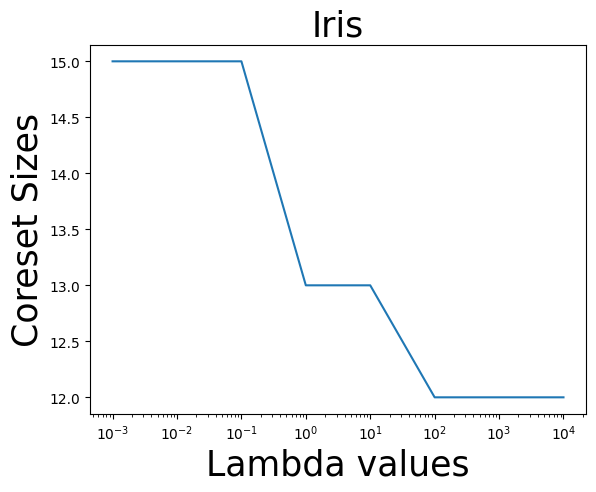}
\includegraphics[width=0.25\linewidth]{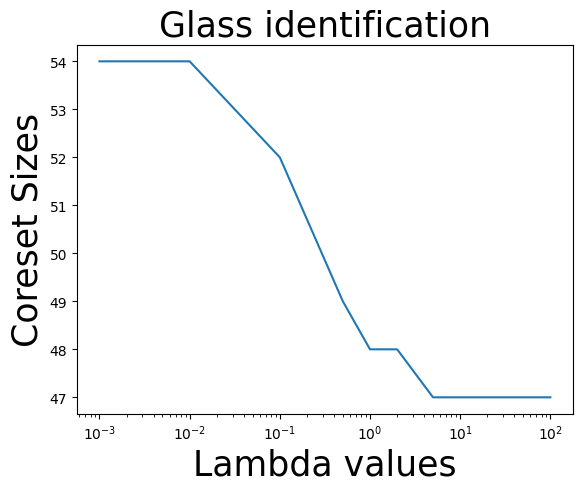}\includegraphics[width=0.25\linewidth]{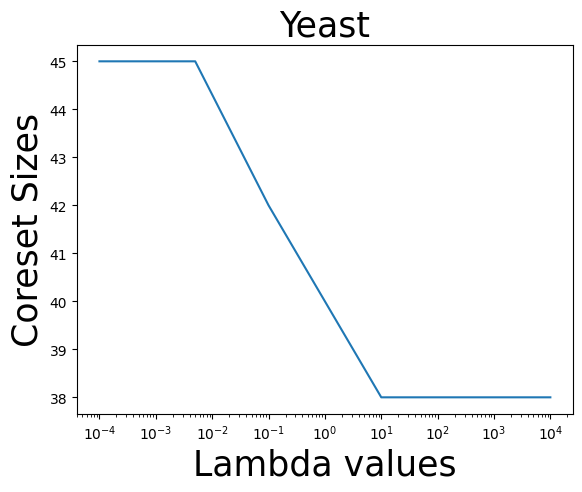}}}
\caption{Accurate Coreset Size vs Regularization Parameter $\lambda$}{\label{fig:size}}
\end{figure*}

For the empirical evaluation of ridge regression, we have used the 1) Breast cancer, 2) Iris, 3) Glass identification and 4) yeast dataset. We have run a simple variant of the algorithm \ref{alg:ridge} on all the datasets for different $\lambda$ values. We use the $\mathrm{Kernelization}$ defined in \eqref{eq:oldridgeKernel}. Let $\{\*c,\*w\}$ be the selected index and its corresponding weights. For every $i \in \*c$ we store $\sqrt{\*w(i)}\cdot\*x_i$ in $\*X_c$ and $\sqrt{\*w(i)}y_i$ in $\*y_c$, if $i \in [n]$. Next, if $i \in [n+1, n+d]$ scale the corresponding standard basis vector with $\*w(i)$ and store it in $\*I_{c}$. Then we train our model as 
\begin{equation}
    \*x^* := \left(\*X_c^T\*X_c + \lambda\*I_{c}\right)^{-1}\*X_c^T\*y_c. \nonumber
\end{equation}
In the figure \ref{fig:size}, it is evident that the size of the accurate coreset decreases (y-axis) with increase in the regularization parameter $\lambda$ (x-axis). This supports our guarantee in Theorem \ref{thm:ridge}. For all the dataset, we obtain optimal RMSE using $\*x^*$ as computed above on $(\*X, \*y)$. 

In table \ref{tab:ridge}, we verify that the RMSE from the model trained on coreset is exactly the same as the RMSE from the model trained on all the four datasets. Our coresets are at most $25\%$ of the training data points among all of them, and we notice a speed up in the training time to $4\times$ times that on the complete data.

\begin{table*}[h]
\caption{Accurate Coreset Results for Regularized Regression} \label{tab:ridge}
\begin{center}
\begin{tabular}{lllllll}
\textbf{Dataset}  &\textbf{Average RMSE} & \textbf{Full Data} & \textbf{Coreset Size} & \textbf{Speed up} &\textbf{Size Reduction}\\
\hline \\
Iris  &  0.339 & $(154 \times 5)$ & $(16 \times 5)$ & $2\times$& $89.61\%$\\
Yeast  & 3.179 & $(1492\times 9)$ & $(46 \times 9)$ & $1.75\times$& $96.9\%$ \\
Breast Cancer  & 0.225 & $(708 \times 10)$ & $(56 \times 10)$ & $4.27\times$ & $92.10\%$\\
Glass Identification  & 0.89 & $(223\times 10)$ & $(56 \times 10)$ & $1.2\times$& $75\%$
\end{tabular}
\end{center}
\end{table*}

\subsection{Gaussian Mixture Models}
Here, we use the 1) Credit card, 2) Covertype and 3) Synthetic dataset. We compute the second order moment $\*T_2$ as described in \cite{anandkumar2014tensor} and compute the whitening matrix $\*M$. Then we multiply $\*M$ with every point in the dataset and also with the standard basis vectors. This reduces the dimension of the space to $\~R^k$ from $\~R^d$ where $d$ is the size of the input space and $k$ is the number of latent variables. Next, we compute the reduced third order moment $\tilde{\@T}_3$ as described in \cite{anandkumar2014tensor}. Finally, we get the accurate coreset of size at most $\binom{k+2}{3}+1$ from our algorithm \ref{alg:lvm}. 

In the table \ref{tab:gmm}, we report the number of latent variables learned, $k$-means clustering loss, with the learned Gaussian parameters on the training dataset. We also report the loss, and as expected, the training loss using the trained model on coreset is the same as the loss using the trained model on the full data. Next, we report the accurate coreset size and speedup in the training time, which is the ratio between the training time on the complete data and training time on the accurate coreset. Finally, we report the reduction in the size of training data. It is important to note that in all the cases, we get a significant advantage in the training time by using coreset. Since the coreset sizes are independent of both $n$ and $d$, so for small datasets (e.g., Credit card), we reduce the training sample size to $0.24\%$ of the original, which gives a significant improvement in the training time. Where as for larger datasets (e.g., Cover type) we reduce the training sample size to $0.02\%$ of the original, which gives a colossal improvement in the training time. 

\begin{table*}[h]
\caption{Accurate Coreset Results for GMM} \label{tab:gmm}
\begin{center}
\begin{tabular}{llllllll}
\textbf{Dataset}  & \textbf{k} &\textbf{Loss} & \textbf{Full Data} & \textbf{Coreset Size} & \textbf{Speed up} &\textbf{Size Reduction}\\
\hline \\
Credit card & 4 & 1025.93 & $(8636 \times 17)$ & $(21 \times 17)$ & $15\times$ & $99.76\%$\\
Synthetic & 3 & 101541.14 & $(10,000 \times 30)$ & $(11 \times 30)$ & $29.25\times$& $99.89\%$\\
Cover type & 4 & 4082.34 & $(101,012 \times 54)$ & $(21 \times 54)$ & $643.8\times$& $99.98\%$
\end{tabular}
\end{center}
\end{table*}



\subsection{Topic Modeling}
Here, we have considered the Research Article dataset, which consists of $20k$ article titles and their abstracts. After cleaning the dataset, we reduce our vocabulary size to $55$. We randomly sample $1k$ documents out of $20k$ and run the single Topic Modeling version of the Algorithm \ref{alg:lvm} for $k=3$. As mentioned in the main paper, even though the original points are in $\~R^{55}$, the coreset size is just $10$, i.e., $\binom{k+2}{3}$. As a result, we improve our learning time by $10$ times compared to learning the latent variable on the complete data of $1k$ documents. 

In the following table we report our anecdotal evidence of three topics learned from the coreset. We identify the top five words from every estimated topic from the coreset. We name the topics based on these top five occurring words in it. We do verify that the topics learnt from the coreset are exactly the same as the topics learnt from the full data. Our running time improves due to fast computation of the tensor $\tilde{\@T}_3$. Using coreset, it takes $0.47$ milliseconds, and from the complete data, it takes $10$ milliseconds. 

\begin{table*}[h]
\caption{Anecdotal Evidence} \label{tab:topic}
\begin{center}
\begin{tabular}{llllllll}
\textbf{Physics} &\textbf{Maths} & \textbf{Computer Science} \\
\hline \\
magnetic & prove&neural \\
spin & finite & training \\
field & multi & deep \\
phase & result & performance \\
quantum & dimensional & optimization 
\end{tabular}
\end{center}
\end{table*}

\section{Conclusion}
In this paper we present a unified framework for constructing accurate coresets for general machine learning models using  $\mathrm{Kernelization}$ technique. Here we presented novel $\mathrm{Kernelization}$ for general problems such as $\ell_p$ regularized $\ell_p$ regression and also for a wide range of latent variable models. Using these we presented algorithms for constructing accurate coresets for the above mentioned problems. 

%% file: section/appendix.tex
\onecolumn
\section{Missing Proofs}

\subsection{Carathéodory's Theorem}
\begin{theorema}[Carathéodory's Theorem]
    Let $\*P$ be a set of $n$ points in $\~R^d$ and it spans $k$-dimensional space. If $\*x$ is a point inside the convex hull of $\*P$, then $\*x$ is also in the convex hull of at most $k+1$ weighted points in $\*P$.
\end{theorema}

\begin{proof}
Let $\*P \in \~R^{n \times d}$ be a set of $n$ points in $\~R^d$, that spans a $k$-dimensional space. So, $\mathrm{rank}(\*P) = k < d$. Let $\*x$ be a point in the convex hull of $\*P$. We can express $\*x$ as a convex combination of points in $\*P$ as $\*x = \sum_{i=1}^{n}\alpha_i \*p_i$, where $\sum_{i=1}^{n} \alpha_i = 1$, and for every $i \in [n], \alpha_i \geq 0$ and $\*p_i \in \*P$. 

Now, let's consider the rank of the space. By the rank-nullity theorem \cite{meyer2023matrix}, we have, $\mathrm{rank}(\*P) + \mathrm{nullity}(\*P) = d$.

If $n > k+1$, then $\mathrm{nullity}(A) = d - \mathrm{rank}(A) > 0$. This implies there exists a non-zero vector $\beta\in \~R^n$ in the null space of $\*P$ such that, $\sum_{i=1}^n \beta_i \*p_i = 0$. 

We can use this to show that the points are linearly dependent. Let's construct a linear combination of the points that equals zero:
\[\sum_{i=1}^n \beta_i (\*p_i - \*p_1) = 0\]
This is equivalent to:
\[\sum_{i=2}^n \beta_i (\*p_i - \*p_1) = 0\]
which is the same form as in the convex hull proof.

Now, we can follow the same steps as in the convex hull proof to reduce the number of points in the convex combination. Define:
\[J = \{i \in \{1,2,\ldots,n\}: \beta_i > 0\}\]
Since $\sum_{i=1}^n \beta_i = 0$ (because $\sum_{i=1}^n \beta_i \*p_i = 0$ and the $\*p_i$ are linearly independent), $J$ is not empty. Define:
\[a = \max_{i \in J} \frac{\alpha_i}{\beta_i}\]
Then we can write:
\[x = \sum_{i=1}^n (\alpha_i - a\beta_i)\*p_i\]
This is a convex combination with at least one zero coefficient. Therefore, we can express $\*x$ as a convex combination of $n-1$ points. We define our new set of $n-1$ weighted points as $\*P$.

We can repeat this process until $\mathrm{nullity}(\*P) = d - \mathrm{rank}(\*P) = 0$.
In $\~R^d$, $\mathrm{nullity}(\*P) = d - \mathrm{rank}(\*P) = 0$ holds true if and only if the number of rows of $\*P$ is less than or equal to $k+1$, where $k$ is the dimension of the space in which the matrix operates.

Hence, for more than $k+1$, points in the space of rank, $k$ are linearly dependent. A point in the convex hull can be expressed as a convex combination of, at most, $k+1$ of these points.
\end{proof}

\subsection{Accurate Coreset Fast Caratheodory}
In all our Algorithms in the main paper, we call the following algorithm \ref{alg:fast_caratheodory} for getting the indices and their weights of the points that are selected as accurate coreset. As we consider our input points are unweighted, each point has equal weights of $1/n$, where $n$ is the number of input points. This is the Fast Caratheodory's algorithm from \cite{maalouf2019fast}. Here, we state the Algorithm \ref{alg:fast_caratheodory} and its subroutine \ref{alg:caratheodory} for completeness.

\begin{algorithm}[htb!]
\caption{Accurate Coreset}\label{alg:fast_caratheodory}
\KwIn{$\*P$ A weighted set of $n$ points in $\mathbb{R}^d$, $\*u$: A weight function $\*u:\*P\to [0,\infty)$, $k > \~R^{+}$ \tcp*[r]{Input}}
\KwOut{$(\*c, \*w)$ \tcp*[r]{Accurate Coreset index and their weights}}

$\*P \gets \*P \setminus \{\*p \in \*P \mid \*u(\*p) = 0\}$ \tcp*[r]{Remove all points with zero weight}

\If{$|\*P| \leq d + 1$}{
    \Return $(\*P, \*u)$ \tcp*[r]{Return if the set is already small enough}
}

Partition $\*P$ into $k$ disjoint subsets $\{\*P_1, \dots, \*P_k\}$, each containing at most $\lceil n/k \rceil$ points \tcp*[r]{Partition the set $\*P$}

\For{$i = 1$ to $k$}{
    $\mu_i \gets \frac{1}{\sum_{\*q \in \*P_i} \*u(\*q)} \sum_{\*p \in \*P_i} \*u(\*p) \cdot \*p$ \tcp*[r]{Compute weighted mean of cluster}
    $\*u'(\mu_i) \gets \sum_{\*p \in \*P_i} \*u(\*p)$ \tcp*[r]{Assign weight to the cluster}
}

$(\tilde{\mu}, \tilde{\*w}) \gets \text{Carathéodory}(\{\mu_1, \dots, \mu_k\}, \*u')$ \tcp*[r]{Apply Carathéodory algorithm on cluster means}

$\*C \gets \bigcup_{\mu_i \in \tilde{\mu}} \*P_i$ \tcp*[r]{Select clusters corresponding to the chosen means}

\For{each $\mu_i \in \tilde{\mu}$ and $\*p \in \*P_i$}{
    $\*w(\*p) \gets \frac{\tilde{\*w}(\mu_i) \*u(\*p)}{\sum_{\*q \in \*P_i} \*u(\*q)}$ \tcp*[r]{Assign new weight for each point in $C$}
}

$(\*C, \*w) \gets \text{Accurate Coreset}(\*C, \*w, k)$ \tcp*[r]{Recursive call}
Compute $(\*c, \*w)$ from $\*C,\*w$ and $\*P$ \tcp*[r]{Indices of selected points in $\*C$ from $\*P$ and their weights}
\Return $(\*c, \*w)$ 
\end{algorithm}

\begin{algorithm}[htb!]
\caption{Carathéodory$(P,w)$}\label{alg:caratheodory}
\KwIn{$\*P$: A weighted set of $n$ points in $\mathbb{R}^d$, where $\*P = \{\*p_1, \dots, \*p_n\}$, $\*u: \*P \to [0, 1]$ \tcp*[r]{Input points}}
\KwOut{A weighted set $(\*S, \*w)$ such that $\*S \subseteq \*P$, $|\*S| \leq d+1$, $\*w: \*S \to [0, 1]$, $\sum_{\*s \in \*S} \*w(\*s) \*s = \sum_{\*p \in \*P} \*u(\*p) \*p$}

\If{$n \leq d+1$}{
    \Return $(\*P, \*u)$ \tcp*[r]{Return if the set is already small enough}
}

\For{$i = 2$ to $n$}{
    $\*a_i \gets \*p_i - \*p_1$ \tcp*[r]{Compute the differences between the points}
}

$\*A \gets (\*a_2 \mid \cdots \mid \*a_n)$ \tcp*[r]{Create the matrix $\*A \in \mathbb{R}^{d \times (n-1)}$}

Compute $\*v = (v_2, \cdots, v_n)^T \neq 0$ such that $\*A \*v = 0$ \tcp*[r]{Find a non-zero vector $\*v$ such that $\*A \*v = 0$}

$\displaystyle v_1 \gets -\sum_{i=2}^n v_i$ \tcp*[r]{Compute $v_1$ from $v_2, \cdots, v_n$}

$\displaystyle \alpha \gets \min\left\{\frac{\*u(\*p_i)}{v_i} \mid i = 1, \cdots, n \text{ and } v_i > 0\right\}$ \tcp*[r]{Compute the step size $\alpha$}

$\*w(\*p_i) \gets \*u(\*p_i) - \alpha v_i$ for every $i = 1, \cdots, n$ such that $\*w(\*p_i) > 0$ \tcp*[r]{Update weights of the points}

$\*S \gets \{\*p_i \mid i = 1, \cdots, n \text{ and } \*w(\*p_i) > 0\}$ \tcp*[r]{Define the new set $\*S$}

\If{$|\*S| > d+1$}{
    $(\*S, \*w) \gets \text{Carathéodory}(\*S, \*w)$ \tcp*[r]{Recursive call to reduce the size of the set}
}

\Return $(\*S, \*w)$ \tcp*[r]{Return the coreset and its weights}
\end{algorithm}